\documentclass[twoside,leqno,twocolumn]{article}

\usepackage[letterpaper]{geometry}

\usepackage{ltexpprt}
\usepackage{graphicx}

\usepackage{amsfonts}
\usepackage{amsmath}
\usepackage{array}
\usepackage{booktabs}
\usepackage{bbm}
\usepackage{bm}
\usepackage{caption}
\usepackage{color}
\usepackage[pdftitle={ourtitle}, hidelinks, hyperfootnotes = false]{hyperref}
\usepackage{multirow}
\usepackage{xspace}
\usepackage{subfigure}
\usepackage[ruled, vlined, nofillcomment, linesnumbered, algo2e]{algorithm2e}

\newcommand{\aka}{\emph{a.k.a.,}\xspace}
\newcommand{\eg}{\emph{e.g.,}\xspace}

\newcommand{\ignore}[1]{}
\newcommand{\paratitle}[1]{\vspace{1.5ex}\noindent\textbf{#1}}

\begin{document}

\title{\Large Neural Graph Matching for Pre-training Graph Neural Networks}
\author{
  Yupeng Hou \thanks{Work done during internship at Ant Group.} \thanks{Gaoling School of Artificial Intelligence, Renmin University of China. $\{$houyupeng,jrwen$\}$@ruc.edu.cn, batmanfly@gmail.com} \thanks{Beijing Key Laboratory of Big Data Management and Analysis Methods.}
\and Binbin Hu \thanks{Ant Group. $\{$bin.hbb,lingyao.zzq,jun.zhoujun$\}$@antfin.com}
\and Wayne Xin Zhao \footnotemark[2] \footnotemark[3] \thanks{Corresponding author.}
\and Zhiqiang Zhang \footnotemark[4]
\and 
\and Jun Zhou \footnotemark[4]
\and Ji-Rong Wen \footnotemark[2] \footnotemark[3] \thanks{School of Information, Renmin University of China.}
}

\date{}

\maketitle

\fancyfoot[R]{\scriptsize{Copyright © 2022 by SIAM\\Unauthorized reproduction of this article is prohibited.}}

\begin{abstract}
  \small
  Recently, 
  graph neural networks (GNNs) have been shown powerful capacity at modeling structural data.
  However, when adapted to downstream tasks, it usually requires abundant task-specific labeled data, which can be extremely scarce in practice.
  A promising solution to  data scarcity  is to pre-train a transferable and expressive GNN model
  on large amounts of unlabeled graphs or coarse-grained labeled graphs.
  Then the pre-trained GNN is fine-tuned on downstream datasets with task-specific fine-grained labels.

  \small
  In this paper, we present a novel \textbf{G}raph \textbf{M}atching based GNN \textbf{P}re-\textbf{T}raining framework,
  called \textbf{GMPT}.
  Focusing on a pair of graphs,
  we propose to learn structural correspondences between them via neural graph matching, consisting of both intra-graph message passing and inter-graph message passing. In this way, we can learn adaptive representations for a given graph when paired with different graphs, and both node- and graph-level characteristics are naturally considered in a single pre-training task.
  The proposed method can be applied to fully self-supervised pre-training and coarse-grained supervised pre-training.
  We further propose an approximate contrastive training strategy to significantly reduce time/memory consumption.
  Extensive experiments on multi-domain, out-of-distribution benchmarks have demonstrated the effectiveness of our approach.
  The code is available at: \url{https://github.com/RUCAIBox/GMPT}.
\end{abstract}

\section{Introduction}\label{sec:intro}

In the past few years, graph neural networks (GNNs) have emerged as
a powerful technical approach for graph representation learning~\cite{kipf2017semi,hamilton2017inductive}.
By leveraging graph structure as well as node and edge features, GNNs can effectively learn low-dimensional representation vectors for each node or the entire graph.
However, to apply GNNs to downstream applications, it
usually requires abundant task-specific labeled data,
which can be extremely scarce in practice.
To alleviate the data scarcity issue~\cite{pan2010survey}, pre-training GNNs~\cite{hu2020strategies,hu2020gpt} has been proposed as a promising solution. 
It first learns a transferable and expressive GNN
on a large number of unlabeled graphs or coarse-grained labeled graphs.
Then, the pre-trained GNN is fine-tuned on downstream datasets with task-specific labels.

For GNN pre-training, existing studies mostly focus on the design of suitable tasks,
such as graph structure reconstruction~\cite{hamilton2017inductive,hu2020strategies,hu2020gpt},
mutual information maximization~\cite{velickovic2019deep,you2020graph,qiu2020gcc}
and properties prediction~\cite{hu2020strategies}.
Generally, these tasks can be classified into two main categories:
(1) \emph{node-level tasks} utilize  node representations to predict the localized properties in the graph (\eg  link prediction);
(2) \emph{graph-level tasks} focus on the entire graph and learn  graph representations when designing the globalized optimization goal (\eg graph's property prediction).

\begin{figure}[t]
  \centering
  \includegraphics[width=0.4\textwidth]{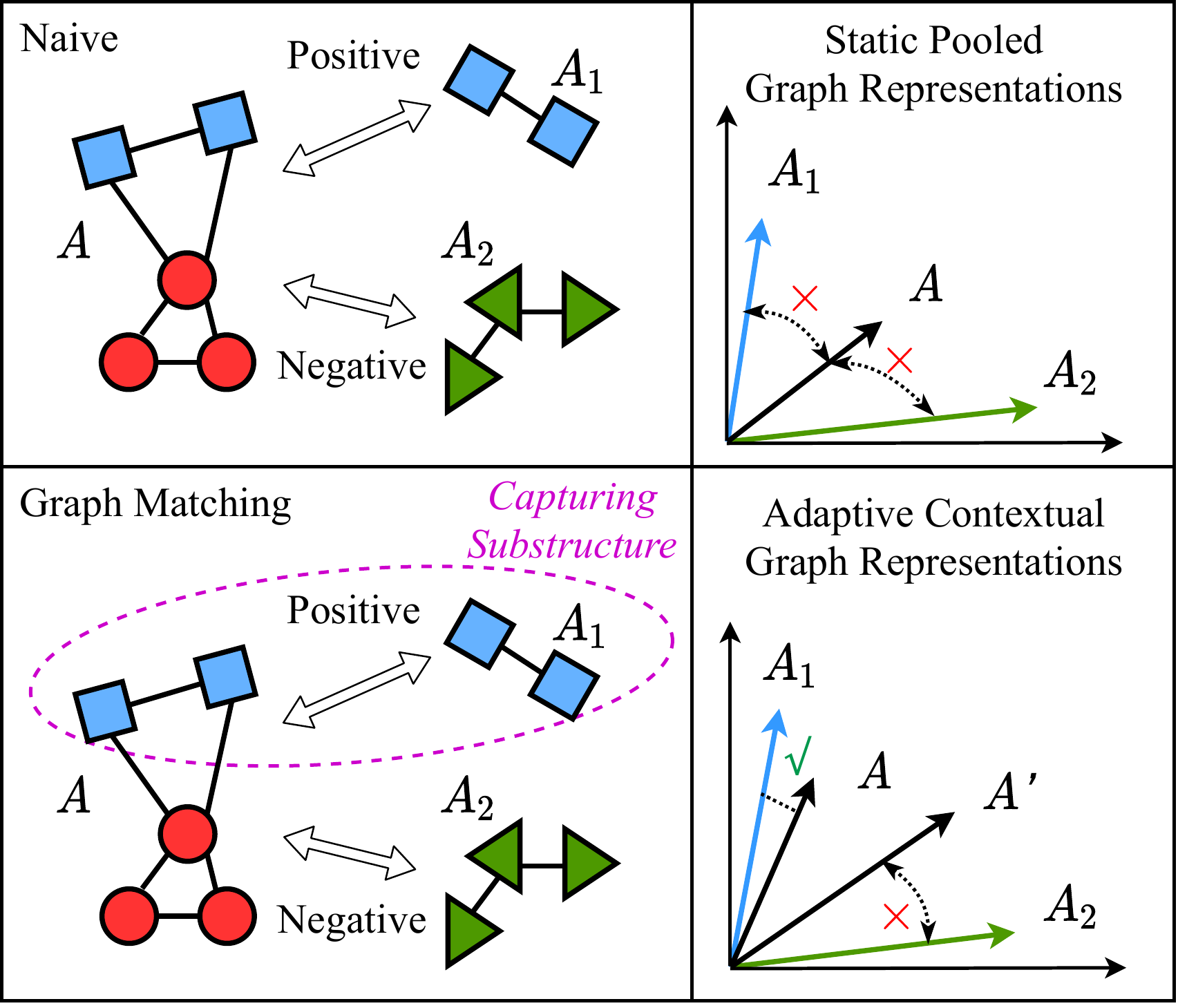}
  \caption{An example of neural graph matching and comparison with existing studies of static graph representations.}
  \label{fig:gm}
\end{figure}

Given the two kinds of GNN pre-training tasks, it is essential to combine node- and graph-level optimization goals~\cite{hu2020strategies}, since they  capture the graph characteristics in different views.
Existing approaches  
either adopt a two-stage approach arranging multi-level pre-training tasks sequentially~\cite{hu2020strategies},
or frame them in a multi-task learning manner~\cite{lu2021learning}.
In this way, each individual pre-training task is not aware of all the optimization goals at different levels, which might result in locally optimal representations \emph{w.r.t.} some specific level (\eg node- or graph-level).
Ideally, a good pre-training task can capture node- and graph-level characteristics simultaneously in order to derive more comprehensive node (and graph) representations.

For this purpose, we attempt to design new GNN pre-training tasks that are able to learn node- and graph-level graph semantics \emph{in one single pre-training task}.
Our solution is based on \emph{neural graph matching}~\cite{wang2019learning,li2019graph,xu2019cross}, a neural approach to learning structural correspondence among graphs. We present an illustrative example of our idea in Figure~\ref{fig:gm}.
At each time,  a pair of two associated graphs (\eg with the same labels or augmented graphs) are given, and we evaluate whether the two graphs have similar structural properties. As a major advantage, neural graph matching 
naturally combines node-level correspondence (\eg $v_1$ to $v_2$) and graph-level properties (\eg whether containing shared substructure)
when establishing their correspondence. 
That is the major reason why we adopt it as the GNN pre-training task.
Another merit of this approach is that a graph will correspond to different representations when paired with different graphs. As shown in this example, we will derive different representations for graph $A$ when paired with graph $A_1$ or $A_2$, since neural graph matching will enforce one graph to refer to another graph's information when learning graph representations. Therefore, we call the learned representations \emph{adaptive graph representations}.
As a comparison, existing graph-level pre-training tasks usually adopt static graph representations.

To this end, in this paper, we propose a novel \textbf{G}raph \textbf{M}atching based GNN \textbf{P}re-\textbf{T}raining method, named as \textbf{GMPT}.
The key contribution lies in a neural graph matching module, where we pair two associated graphs  as input at each time. 
To learn structural  correspondences,
 we perform intra-graph as well as inter-graph message passing. In this way, the representations of a given graph are learned by referring to another paired graph, which derives adaptive graph representations.
Such a method can capture both node- and graph-level characteristics when learning the graph representations. 
The proposed method can be applied to both
fully self-supervised and coarse-grained supervised pre-training settings.
In self-supervised setting, GNNs are optimized by a graph matching-based contrastive loss.
To accelerate the learning of graph pairs during pre-training, we further proposed an \emph{approximate contrastive training} strategy to significantly reduce the time/memory consumption, without loss of accuracy.
While in supervised setting,
we design different supervised tasks according to different coarse-grained labels.

In summarization, we design a new GNN pre-training task based on neural graph matching,
devoted to adaptive graph representation learning by modeling both node- and graph-level characteristics in a single pre-training task.
We also propose an approximate contrastive training strategy to reduce the time/memory consumption. Extensive experiments on public out-of-distribution benchmarks from multiple domains on various GNN architectures have demonstrated the effectiveness of our approach.

\begin{figure*}[ht]
    \centering
    \includegraphics[width=0.85\textwidth]{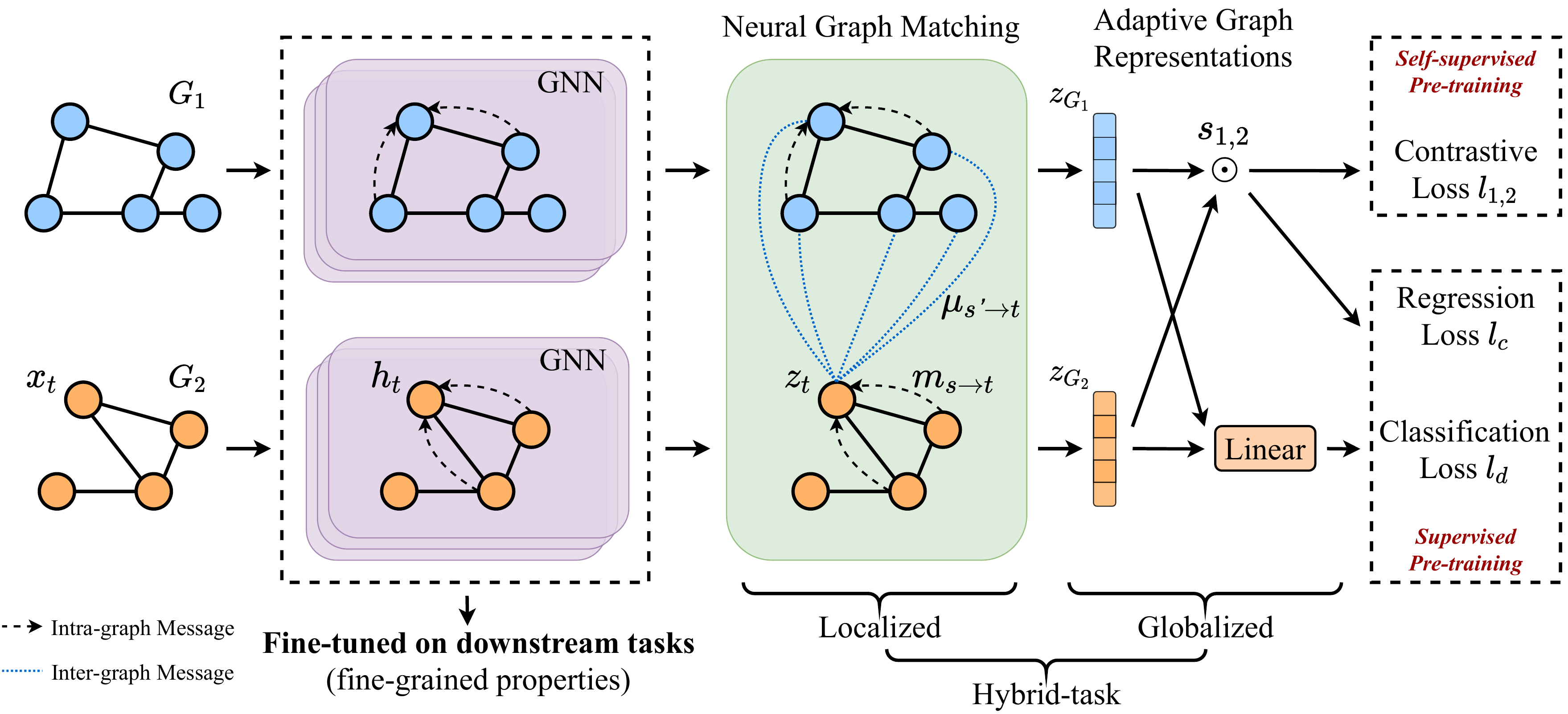}
    \caption{Overall framework of our proposed graph matching-based GNN pre-training methods.}
    \label{fig:overall}
  \end{figure*}

\section{Related Work}
In this section, we review the most related work about pre-training graph neural networks and graph matching.

\subsection{Pre-training Graph Neural Networks}\hfill\\
Though graph neural networks are powerful tools to characterize graph-structured data, they heavily rely on fine-grained domain-specific labels while training, which is usually scarce and difficult to obtain.
To alleviate the above issues, pre-training for graph neural networks has drawn much attention recently, which empowers GNNs to capture the structural and semantic information of the input unlabeled graphs (or with few coarse-grained labels), followed by several fine-tuning steps on the downstream tasks of interest. Obviously, developing effective supervised (self-supervised) signals to guide GNNs to exploit structural and semantic properties on original graphs is at the heart.
Generally, existing designed supervised signals can be classified into two main categories.
The first is called \emph{node-level tasks}, which aims at predicting localized properties utilizing node representations,
such as graph structure reconstruction~\cite{hamilton2017inductive,hu2020strategies,hu2020gpt,lu2021learning,li2021explicit}, localized attribute prediction~\cite{hu2020strategies,hu2020gpt} and node representation recovery~\cite{hao2021pre}.
Another is called \emph{graph-level tasks}, which defines globalized optimization goal for the entire graph,
such as graph property prediction~\cite{hu2020strategies,li20221pairwise} and mutual information maximization~\cite{velickovic2019deep,you2020graph,qiu2020gcc,lu2021learning,xu2021log}.
Our proposed framework differs from the above approaches in the following two aspects: learning node- and graph-level graph semantics \emph{in one single pre-training task} and \emph{adaptive graph representations}.

\subsection{Graph Matching}\hfill\\
Graph matching refers to establishing node correspondences between two (or among multiple) graphs~\cite{caetano2009learning},
such that the similarity between the matched graphs is maximized.
Some researches focus on the accuracy of the node correspondence,
and regard graph matching as a quadratic assignment programming (QAP) problem~\cite{loiola2007survey}, which is NP-complete.
Thus, researchers mainly employ approximate techniques to seek inexact solutions, such as spectral approximation~\cite{leordeanu2005spectral},
double-stochastic approximation~\cite{gold1996graduated},
and learning-based approaches~\cite{caetano2009learning,zanfir2018deep,wang2019learning}.
While others care about the similarity calculation between graphs.
Early efforts are mainly based on heuristic rules, such as minimal graph edit distance~\cite{willett1998chemical}
and graph kernel methods~\cite{kashima2003marginalized,vishwanathan2010graph}.
With the development of GNNs, recent work leverage message passing techniques to explore neural-based graph matching~\cite{li2019graph,xu2019cross}.
In this work, we apply neural graph matching for GNN pre-training, to learn adaptive graph representations
and encourage GNNs to integrate localized and globalized domain-specific features.

\section{The Proposed Method}

In this section, we first introduce the notation and problem definition.
Then we present the proposed graph matching based GNN pre-training method \textbf{GMPT}
for both self-supervised setting and coarse-grained supervised setting.
Our approach takes pairs of graphs as input.
With a carefully designed cross-graph message passing mechanism,
one graph can be adaptively encoded into different graph representations when paired with different graphs.
Figure~\ref{fig:overall} presents the overall architecture of our proposed framework.

\subsection{Notation and Problem Definition}\hfill\\
A graph can be represented as $G=(\mathcal{V},\mathcal{E}, \bm{X}, \bm{E})$,
where $\mathcal{V} = \{v_1, v_2, \ldots, v_{|\mathcal{V}|}\}$ denotes the node set,
$\mathcal{E} \subseteq \mathcal{V} \times \mathcal{V}$ denotes the edge set,
$\bm{X}\in \mathbb{R}^{|\mathcal{V}|\times d_{v}}$
and $\bm{E} \in \mathbb{R}^{|\mathcal{E}|\times d_{e}}$ represent the $d_v$- and $d_e$-dimensional attribute matrix for nodes and edges, respectively.
Furthermore, each graph is possibly associated with some label $y$ from a label set $\mathcal{Y}$. 
Given a set of graphs with labels $\{(G_1, y_1),(G_2, y_2), \ldots, (G_N, y_N)\}$,
graph neural networks (GNNs) leverage graph structure as well as node and edge features
to learn a representation vector for
the entire graph $\bm{h}_G$,
and further utilize $\bm{h}_G$ to predict the corresponding label $y$.
A detailed preliminary of GNNs is provided in Supplementary Material~S.1\footnote{Please refer to: \url{https://github.com/RUCAIBox/GMPT/blob/main/paper/supplementary_material_gmpt.pdf}}.

In this work, we focus on pre-training GNNs: GNNs that are initialized with pre-trained parameters
are fine-tuned according to various downstream tasks.
Given the defined graph learning task, we consider two kinds of GNN pre-training paradigms based on whether graph labels are used or not during pre-training: \emph{self-supervised setting} (without graph labels) and \emph{supervised setting} (with graph labels).

\subsection{Self-supervised Pre-training}\label{sec:gmcl}\hfill\\
We have no available labeled data for pre-training in self-supervised setting. The pre-training task is to evaluate whether a pair of augmented graph views are generated based on the same graph. We adopt contrastive learning  to construct the learning objective and name our approach in this setting
as \textbf{GMPT-CL}.

\subsubsection{Graph Representation Learning via Graph Matching}\label{sec:context_rep} 
We first present how to learn graph representations via graph matching. 

\paratitle{Graph augmentation and encoding}. Given a list of $n$ graph examples,
we first apply stochastic data augmentation to transform any given graph example
into two correlated views randomly ($2n$ views in total).
We consider various augmentation techniques, including  node/edge perturbation~\cite{you2020graph},
subgraph sampling~\cite{you2020graph},
diffusion~\cite{hassani2020contrastive}, and adaptive methods~\cite{zhu2021graph}.
The selection of graph augmentation techniques depends on the actual data domain~\cite{you2020graph}.
To pre-train an expressive GNN encoder,
we consider whether a pair of graph views denoted by $\tilde{G}_1$
and $\tilde{G}_2$ are matched or not based on their graph representations. 
Specially, we first apply the GNN encoder to obtain node representations in the two graph views. 
Let $\bm{h}^{(1)}_s$ and $\bm{h}^{(2)}_t$ denote the representations of node $s$ from $\tilde{G}_1$ and node $t$ from $\tilde{G}_2$, respectively.

\paratitle{Neural graph matching}.  
Following recent progress in neural graph matching~\cite{li2019graph,wang2019learning},  we  
incorporate message passing within a graph (called \emph{intra-graph message})
and between a pair of graphs (called \emph{inter-graph messages}).
Given a intra-graph node pair $\langle s, t\rangle$ and a inter-graph node pair $\langle s', t'\rangle$, we define the two kinds of message passing mechanisms formally as:
\begin{align}
  \bm{m}_{s \rightarrow t}&= \operatorname{MSG}_{\operatorname{intra}}\left(\bm{h}_{s}^{(1)}, \bm{h}_{t}^{(1)}, \bm{e}_{s t}\right),\label{eq:intra-graph-mes} \\
  \bm{\mu}_{s' \rightarrow t'}&= \operatorname{MSG}_{\operatorname{inter}}\left(\bm{h}_{s'}^{(1)}, \bm{h}_{t'}^{(2)}\right),\label{eq:inter-graph-mes}
\end{align}
where $\bm{m}_{s \rightarrow t}$ and $\bm{\mu}_{s' \rightarrow t'}$ are
intra-graph and inter-graph messages, respectively.
Intra-graph message passing can be defined in a similar way following standard GNN architectures, like GIN~\cite{xu2019how}.
While
for $\operatorname{MSG}_{\operatorname{inter}}$, we adopt a cross-graph attention mechanism as:
\begin{align*}
  \bm{\mu}_{s' \rightarrow t'} =%
    &\ a_{s' \rightarrow t'}\cdot\bm{h}_{s'}^{(1)},
\end{align*}
where $a_{s' \rightarrow t'}=\frac{\exp(\operatorname{sim}(\bm{h}_{s'}^{(1)}, \bm{h}_{t'}^{(2)}))}{\sum_{k \in \tilde{G}_2 } \exp(\operatorname{sim}(\bm{h}_{s'}^{(1)}, \bm{h}_{k}^{(2)}))}$ and 
$\operatorname{sim}(\cdot)$ is a similarity function,
such as dot product and cosine similarity.
The above attention mechanism allows an adaptive message exchange between the paired graphs.
Intuitively, messages passed between similar substructures of graphs will have higher attention weights.
Here, we normalize the attention weights of messages from the same \emph{source} node,
which means $\sum_{t'} a_{s' \rightarrow t'} = 1$.
Besides, it is also optional to normalize the attention weights of messages to the same \emph{target} nodes ($\sum_{s'} a_{s'\rightarrow t'} = 1$).

\paratitle{Match enhanced graph representations.} After passing intra-graph messages from nodes' neighbors (denoted by $\mathcal{N}_{\text{intra}}$)
and inter-graph messages from all the nodes of another graph (denoted by $\mathcal{N}_{\text{inter}}$),
we aggregate the messages together and update to obtain the contextual node features $\bm{Z}$.
For a node $t$, we update its original representation $\bm{h}_{t}$ as:
\begin{align}
  \bm{z}_{t}&= \operatorname{Update}\left(\bm{h}_{t}, \sum_{s\in \mathcal{N}_{\text{intra}}} \bm{m}_{s \rightarrow t}, \sum_{s'\in \mathcal{N}_{\text{inter}}} \bm{\mu}_{s' \rightarrow t}\right),\label{eq:update}
\end{align}
where we use the sum operation for aggregation.
Finally, we obtain the entire graph’s adaptive representation $\bm{z}_G$ by
employing a permutation-invariant function $\operatorname{READOUT}$ to pool contextual node features:
\begin{align}
  \bm{z}_{G} &= \operatorname{READOUT}\left(\{\bm{z}_{v}|v\in\mathcal{V}\}\right).\label{eq:z12}
\end{align}
Note that when involved in different pairs, a given graph will correspond to different representations in our approach. 
It is a key merit for subsequent pre-training tasks, since it can adaptively capture structural correspondences instead of using static graph representations as in previous studies~\cite{hu2020strategies,you2020graph}. 

\subsubsection{Contrastive Learning with Adaptive Graph Representations}\label{sec:hybrid_level_rep}

Contrastive learning is a commonly used technique to learn with augmented graph views in pairs~\cite{you2020graph}. It aims to increase 
similarity scores for positive pairs and decrease similarity scores for negative pairs.
However, existing graph contrastive learning methods mainly adopt static graph representations~\cite{you2020graph,hassani2020contrastive,zhu2021graph}, where node-level interaction across graphs is not explicitly modeled 
in this process. As a comparison, given a pair of graph views, we first apply the neural graph matching technique (Section~\ref{sec:context_rep}) to characterize inter-graph interaction, and then construct the  contrastive loss based on the adaptive graph representations. 

Formally, given a positive pair $(\tilde{G_i}, \tilde{G_j})$,
we firstly adaptively encode them into contextual graph representations $\bm{z}_{\tilde{G_i}}$ and $\bm{z}_{\tilde{G_j}}$ (Eqn.~\eqref{eq:z12}), 
and then formalize the  contrastive loss below:
\begin{equation}
  \ell_{i, j}=-\log \frac{\exp \left(s_{i, j} / \tau\right)}{\sum_{k \neq i}\exp \left(s_{i, k} / \tau\right)},\label{eq:cl}
\end{equation}
where $ s_{i, j} = \operatorname{sim}(\bm{z}_{\tilde{G_i}}, \bm{z}_{\tilde{G_j}})$ and $\tau$ is a temperature parameter.
In practice, we usually have a batch of graph views, and we enumerate all the possible pairs of graphs in a batch for this loss in the denominator of Eqn.~\eqref{eq:cl}.

Although the above contrastive loss is also defined at the graph level (whether two views are augmented from the same graph),
the derived graph representations $\bm{z}_{\tilde{G_i}}$ and $\bm{z}_{\tilde{G_j}}$ are enhanced with inter-graph node interaction via neural graph matching. 
As such, optimizing $\ell_{i,j}$ will encourage GNNs to capture both node- and graph-level characteristics in graph representations.

\subsubsection{Approximate Contrastive Training}\label{sec:appro}
A major issue with graph matching is it incurs a quadratic  time and space cost in terms of the number of nodes. 
Here, we propose an approximate contrastive training strategy to improve algorithm efficiency. 

\paratitle{Complexity analysis.}
We consider the setting with a mini-batch of $n$ graphs. As mentioned before, we would generate $2n$ augmented graph views for graph matching and  contrastive learning.  
Typically, for a mini-batch of $n$ graphs,
GMPT-CL considers $2n\times2n$ times of graph comparisons (two augmented views each graph) in total.
Each comparison performs a node-to-node similarity calculation (refer to Eqn.~\eqref{eq:inter-graph-mes}),
taking an additional cost of $O(m^2\cdot d)$ time and space, where $m=\sum^{2n}_{i=1}|\mathcal{V}_i|$ denotes the total number of nodes in $2n$ graph views and $d$ is the dimensionality of representation vectors.

\paratitle{Approximate calculation.}
In order to reduce time and memory consumption, the key idea 
is to perform an approximate calculation of the proposed contrastive loss (Eqn.~\eqref{eq:cl}):
we sample $q$  out of $2n$ graph views
to contrast with all the other views ($q \times 2n$ times comparisons totally).
In this way, the additionally expected time complexity can be reduced to $O(\frac{q}{2n}\cdot m^2\cdot d)$.
For a further reduction of space complexity, we adopt the gradient accumulation technique.
For each time of sampling, we perform $1\times 2n$ times of comparison.
After calculating the contrastive loss,
the model backpropagates prediction error and calculates the gradients,
but doesn't update model parameters immediately.
Instead, the gradients are accumulated until all the $q$ samples are calculated.
In this way, the sampled $q$ graphs only require an additional space complexity of $O\left(\frac{1}{2n}\cdot m^2\cdot d\right)$.

\paratitle{Theoretical analysis.}
We provide a theoretical analysis to reveal 
 the connection between GMPT-CL with approximate contrastive training and mutual information maximization.
Firstly, we show that,
\begin{lemma}
  Minimizing Eqn.~\eqref{eq:cl} is equivalent to
  maximizing a lower bound of the mutual information
  between the latent representations of two views of graphs.
\end{lemma}
Furthermore, we show that,
\begin{lemma}
  Optimizing Eqn.~\eqref{eq:cl} with approximate contrastive training algorithm has the same optimization lower bound as originally in expectation.
\end{lemma}
Thus we can see that the proposed approximate contrastive training
fits the formulation of the InfoNCE loss~\cite{oord2018representation,tschannen2020on} in expectation.
The proofs are provided in Supplementary Material~S.3.

Empirically, experiments in Section~\ref{sec:exp-training}
will show that
performances of the fine-tuned GNNs on downstream tasks don't drop
(even improve) with the proposed approximate contrastive training.
The overall pre-training algorithm of GMPT-CL in one mini-batch is provided in Supplementary Material~S.3.

\subsection{Supervised Pre-training}\label{sec:supgm}\hfill\\
Besides a large number of unlabeled graphs for self-supervised pre-training,
sometimes we can obtain graphs with coarse-grained labels.
Different from elaborately created fine-grained labels,
coarse-grained labels have a relatively weak correlation with downstream task goals, but can be obtained in an easier way.
For example, in molecular property prediction,
we can easily collect the properties of molecules that have been experimentally measured so far~\cite{hu2020strategies}.

In supervised pre-training setting, we pre-train GNNs on graphs with coarse-grained labels,
and then the pre-trained GNNs are fine-tuned according to fine-grained labels in downstream tasks.
Note that coarse-grained labels used for supervised pre-training are not the real labels of downstream tasks.
Based on whether coarse-grained labels are continuous or discrete, we propose two variants of GMPT in supervised setting, 
named as \textbf{GMPT-Sup} and \textbf{GMPT-$\text{Sup}_{++}$}, respectively.

\paratitle{Continuous labels.}
Labels with continuous properties can be regarded as real-value vectors.
We assume that similar pairs of graphs also correspond to similar labels. 
Based on this consideration, we propose \textbf{GMPT-Sup} to learn the similarities between graphs via graph matching module,
and then minimize the difference between the learned similarity  and the actual label similarity.
Given a pair of graphs $G_1$ and $G_2$ and their coarse-grained labels $y_1$ and $y_2$,
we firstly obtain their adaptive graph representations $\bm{z}_{G_1}$ and $\bm{z}_{G_2}$ as Eqn.~\eqref{eq:z12}.
Then we define the loss function as $\ell_{c} = \operatorname{MSE}(s_p, s_{g})$,
where $s_p = \operatorname{sim}(\bm{y}_1, \bm{y}_2), s_{g} = \operatorname{sim}(\bm{z}_{G_1},\bm{z}_{G_2})$,
and $\operatorname{MSE}(\cdot,\cdot)$ denotes the standard mean squared error loss.
This loss drives the representation similarity of two graphs to be close to their label similarity. 

\paratitle{Discrete labels.}
For discrete labels, we do not enforce a direct comparison of the two graphs in a pair. Instead, 
we construct a classification-based approach \textbf{GMPT-$\text{Sup}_{++}$} to associated graph representations with suitable coarse-grained labels.
For a given pair of graphs $G_1$ and $G_2$ (encoded by graph matching module into $\bm{z}_{G_1}$ and $\bm{z}_{G_2}$),
we jointly predict the corresponding coarse-grained labels $y_1$ and $y_2$.
The loss for discrete labels $\ell_{d}$ can be calculated as,
\begin{align*}
  \ell_{d} = \sum_{k=1,2}\operatorname{BCE}(\bm{y}_k, \bm{W}_k\cdot\bm{z}_{G_k}+\bm{b}_k),
\end{align*}
where $\operatorname{BCE}(\cdot,\cdot)$ denotes the popular binary cross-entropy loss,
and  $\bm{y}_1$ and $\bm{y}_2$  are vectorized representations of labels $y_1$ and $y_2$, respectively. 
Although the loss of the two graphs in a pair is calculated separately, their representations are obtained from the graph matching module (\eg cross-graph message passing in Eqn.~\eqref{eq:inter-graph-mes}).

\section{Experiments}

In this section, we conduct extensive experiments to verify the effectiveness of our proposed methods in both self-supervised and supervised settings. Moreover, we also give an in-depth analysis of training strategy and transferability.
Additional experiments on key parameters sensitivity analysis are provided in Supplementary Material~S.5.4.

\begin{table}
  \caption{
    Statistics of the datasets.
    \emph{PT} denotes Pre-Training and \emph{FT} denotes Fine-Tuning.
    * denotes the total number of tasks for the eight downstream datasets.
  }
  \label{tb:dataset}
  \centering
  \resizebox{\columnwidth}{!}{
  \begin{tabular}{lcc}
    \toprule
    Dataset & Bio & Chem  \\
    \midrule
    \midrule
    \#(sub)graphs for self-supervised PT & $307$K & $2,000$K \\
    \#(sub)graphs for supervised PT/FT & $88$K & $456$K \\
    \#Coarse-grained labels for PT & $5,000$ & $1,310$  \\
    \#Downstream FT tasks & $40$ & $678$* \\
    \bottomrule
  \end{tabular}
  }
\end{table}

\subsection{Experimental Setup}
\subsubsection{Datasets}
We conduct experiments on two public out-of-distribution (sub)graph classification benchmarks from different domains, namely Bio and Chem~\cite{hu2020strategies}.
We strictly adopt the same way of splitting and pre-processing of these benchmarks as previous work~\cite{hu2020strategies}.
Dataset statistics are summarized in Table~\ref{tb:dataset}.
Detailed descriptions of the datasets are given in Supplementary Material~S.5.1.

\begin{table*}[!t]
  \small
  \caption{
    Evaluation in self-supervised setting.
    We test ROC-AUC (\%) performance using different pre-training methods.
    Besides, performances with different GNN architecture on Bio are also presented.
    The macro-average results over all GNN architectures on Bio, and results of GIN over all subtasks on Chem are also reported.
    In L2P-GNN, GNN is fine-tuned with a parameterized global pooling layer, while others use average pooling.
  }
  \label{tb:overall}
  \centering
  \begin{tabular}{ccccccc}
    \toprule
    \multirow{2}{*}{\textbf{Pre-training methods}} & \multicolumn{5}{c}{\textbf{Bio}} & \multirow{2}{*}{\textbf{Chem}} \\
     & GCN & GraphSAGE & GAT & GIN & Average & \\
    \midrule
    \midrule
    w/o pre-training          & $63.20 \pm 1.00$        & $65.70 \pm 1.20$        & $68.20 \pm 1.10$        & $64.80 \pm 1.00$        & $65.48$   & $67.0$    \\
    Infomax     & $62.83 \pm 1.22$       & $67.21 \pm 1.84$       & $66.94 \pm 2.61$       & $64.10 \pm 1.50$       & $65.27$   & $70.3$   \\
    EdgePred    & $63.18 \pm 1.12$ & $66.05 \pm 0.78$       & $65.72 \pm 1.17$       & $65.70 \pm 1.30$        & $65.16$  & $70.3$    \\
    ContextPred & $62.81 \pm 1.87$        & $66.47 \pm 1.27$        & $67.86 \pm 1.19$        & $65.20 \pm 1.60$        & $65.59$  & $71.1$    \\
    AttrMasking & $62.40 \pm 1.35$        & $63.32 \pm 1.01$        & $61.72 \pm 2.70$        & $64.40 \pm 1.30$        & $62.96$ & $70.9$     \\
    GraphCL     & $67.05 \pm 1.16$        & \textbf{71.53} $\pm 0.46$   & $65.68\pm 3.98$        & $67.88\pm 0.85$        & $68.04$ & $70.8$     \\
    L2P-GNN     & $66.48 \pm 1.59$ & $69.89 \pm 1.63$ & $69.15 \pm 1.86$ & $70.13 \pm 0.95$ & $68.91$ & $70.4$ \\
    \midrule
    GMPT-CL     & \textbf{70.65} $\pm\ 0.53$ & $70.29 \pm 0.21$ & \textbf{71.07} $\pm\ 0.14$ & \textbf{72.53} $\pm\ 0.42$ & \textbf{71.13} & \textbf{71.5} \\
    \bottomrule
  \end{tabular}
\end{table*}

\subsubsection{Baselines}

We compare our pre-training methods with the following representative GNN pre-training methods:\\
$-$ \textbf{Infomax}~\cite{velickovic2019deep} maximizes the mutual information between patch representations and corresponding high-level summaries of graphs.\\
$-$ \textbf{EdgePred}~\cite{hamilton2017inductive} directly predicts the connectivity of node pairs, \aka link prediction task.\\
$-$ \textbf{ContextPred}~\cite{hu2020strategies} uses subgraphs to predict their surrounding graph structures. \\
$-$ \textbf{AttrMasking}~\cite{hu2020strategies} predicts nodes' or edges' attributes, which are randomly masked.\\
$-$ \textbf{GraphCL}~\cite{you2020graph} contrast the static representations of augmented views and judge whether they are generated from the same graph.\\
$-$ \textbf{L2P-GNN}~\cite{lu2021learning} utilizes meta-learning to alleviate the divergence between multi-task pre-training and fine-tuning objectives.\\
$-$ \textbf{PropPred}~\cite{hu2020strategies} predicts the coarse-grained labels of graphs in the pre-training datasets. \\

Among the baselines, PrepPred is designed for supervised pre-training setting, while others are targeted at  self-supervised pre-training setting. Moreover, results of the non-pre-trained model are also reported.

\subsubsection{Parameter Settings}
To enhance the reproducibility, we elaborately present the implementation details as follows.
Detailed hyper-parameters for GNN architecture and training are provided in Supplementary Material~S.5.2.

\paratitle{GNN architecture.}
We mainly experiment on Graph Isomorphism Networks (GINs)~\cite{xu2019how},
the most expressive GNN architecture for graph-level prediction tasks.
We also experiment with other popular architectures: GCN~\cite{kipf2017semi}, GraphSAGE
~\cite{hamilton2017inductive} and GAT~\cite{velickovic2018graph}.
We select the same GNN hyper-parameters as previous works~\cite{hu2020strategies,lu2021learning}.
For the proposed graph matching methods,
we adopt
dot production for the $\operatorname{sim}(\cdot)$ function
and select $\tau=0.07$ in Eqn.~\eqref{eq:cl}. We utilize a multiple layer perceptron (MLP) as function of $\operatorname{Update}$ in Eqn.~\eqref{eq:update}.

\paratitle{Pre-training and fine-tuning settings.}
Results of baselines on different datasets are directly taken if they have been reported literaturely.
For the other method, we pre-train the models with a learning rate of $0.001$,
and fine-tune the GNNs with a learning rate tuned in $\{0.01, 0.001, 0.0001\}$ for all the methods.
We report the ROC-AUC for both datasets.
The downstream experiments are run with $10$ random seeds, and we report the mean and standard deviation of the metrics.

\subsection{Performance Comparison}

\subsubsection{Self-supervised Setting}

Table~\ref{tb:overall} presents the performance comparison
in self-supervised pre-training setting 
between GMPT-CL and the baselines.

The proposed pre-training method GMPT-CL achieves the best performance $72.53\%$
over all the compared methods on Bio.
with the most expressive GNN architecture GIN.
On Chem dataset, we also notice that GMPT-CL gains the best results ($71.5\%$) compared to all the baselines.

Applying GMPT-CL on currently popular GNN architectures, as shown in Table~\ref{tb:overall},
GMPT-CL achieves the best macro-average result ($71.13\%$) over all the compared baseline methods.
In particular,  we can see that GMPT-CL is powerful even with less expressive GNN architectures like GCN or GAT,
which brings $3.60\%$ and $1.92\%$ absolutely gains compared to the best baseline, respectively.

In sum, we make the following observations.

\textbf{(1)} On average, the proposed GMPT-CL yields the best performance on benchmarks of different domains ($72.53\%$ on Bio and $71.5\%$ on Chem).
  As GMPT-CL is a hybrid-level pre-training task, it encourages GNNs to capture both localized and globalized domain-specific semantics.
  Graph matching module of GMPT-CL can generate adaptive graph representations,
  in which shared substructures are enhanced.

\textbf{(2)} Pre-training GNNs with a large amount of unlabeled data is clearly helpful to downstream tasks,
  as GMPT-CL brings $6.69\%$ and $4.5\%$ absolutely gains compared to non-pre-trained models on the macro-average results over datasets of two domains, respectively.

\subsubsection{Supervised Setting}

\begin{table}
  \small
  \caption{
    Evaluation in supervised setting.
    We report ROC-AUC (\%) performance on Bio and Chem using different supervised pre-training methods with GIN.
  }
  \label{tb:sup}
  \centering
  \begin{tabular}{ccc}
    \toprule
    \textbf{Pre-training methods} & \textbf{Bio} & \textbf{Chem} \\
    \midrule
    \midrule
    w/o pre-training                   & $64.8 \pm 1.0$   & $67.0$ \\
    PropPred             & $69.0 \pm 2.4$   & $70.0$ \\
    \midrule
    GMPT-Sup                & \textbf{70.84 $\pm$ 0.59}   & --       \\
    GMPT-$\text{Sup}_{++}$              & $70.73 \pm 0.42$ & \textbf{70.4} \\
    \bottomrule
  \end{tabular}
\end{table}

Table~\ref{tb:sup} presents the performance comparison between GMPT-Sup, GMPT-$\text{Sup}_{++}$, and the baselines (i.e., non-pre-trained GIN and PropPred) in supervised pre-training setting.
Though coarse-grained labels in Bio datasets are multi-hot vectors, we still view them as continuous properties in GMPT-Sup.
While coarse-grained labels in Chem datasets contain plenty of missing values.
As similarity over missing values is hard to define,
GMPT-Sup is not a suitable method for labels with missing values, and we only report the result of GMPT-$\text{Sup}_{++}$ on Chem.

The compared baseline PropPred always outperforms the non-pre-trained method,
reflecting that GNNs pre-trained on coarse-grained labels can characterize domain-specific semantics.
Compared with PropPred, which encodes graphs into static representations with a graph-level pre-training task,
the proposed hybrid-level methods GMPT-Sup and GMPT-$\text{Sup}_{++}$ generate adaptive graph representations (Section~\ref{sec:context_rep}).
Consequently, we can see that the proposed methods achieve the best performances on Bio and Chem, respectively,
demonstrating the effectiveness of our pre-training methods.
Besides, we notice that GMPT-Sup outperforms GMPT-$\text{Sup}_{++}$ slightly on Bio dataset.
A possible reason is that directly predicting every single property of coarse-grained labels (as GMPT-$\text{Sup}_{++}$) may cause overfitting
and limit the transferability of the pre-trained model, especially when coarse-grained labels lack precious domain-specific semantics.

\begin{figure}
  \centering
  \subfigure{\label{fig:sample:auc}\includegraphics[width=0.24\textwidth]{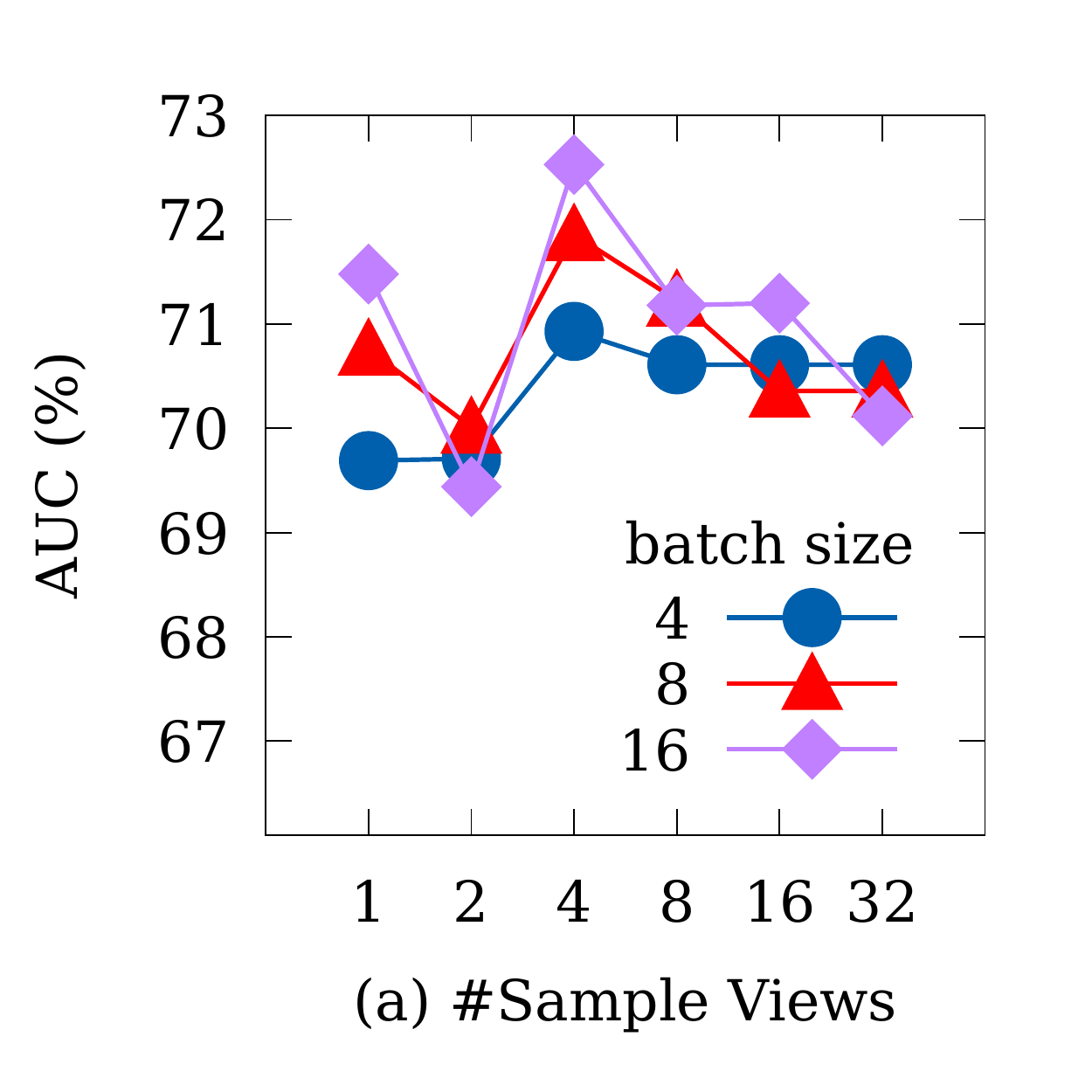}}
  \subfigure{\label{fig:sample:timespace}\includegraphics[width=0.24\textwidth]{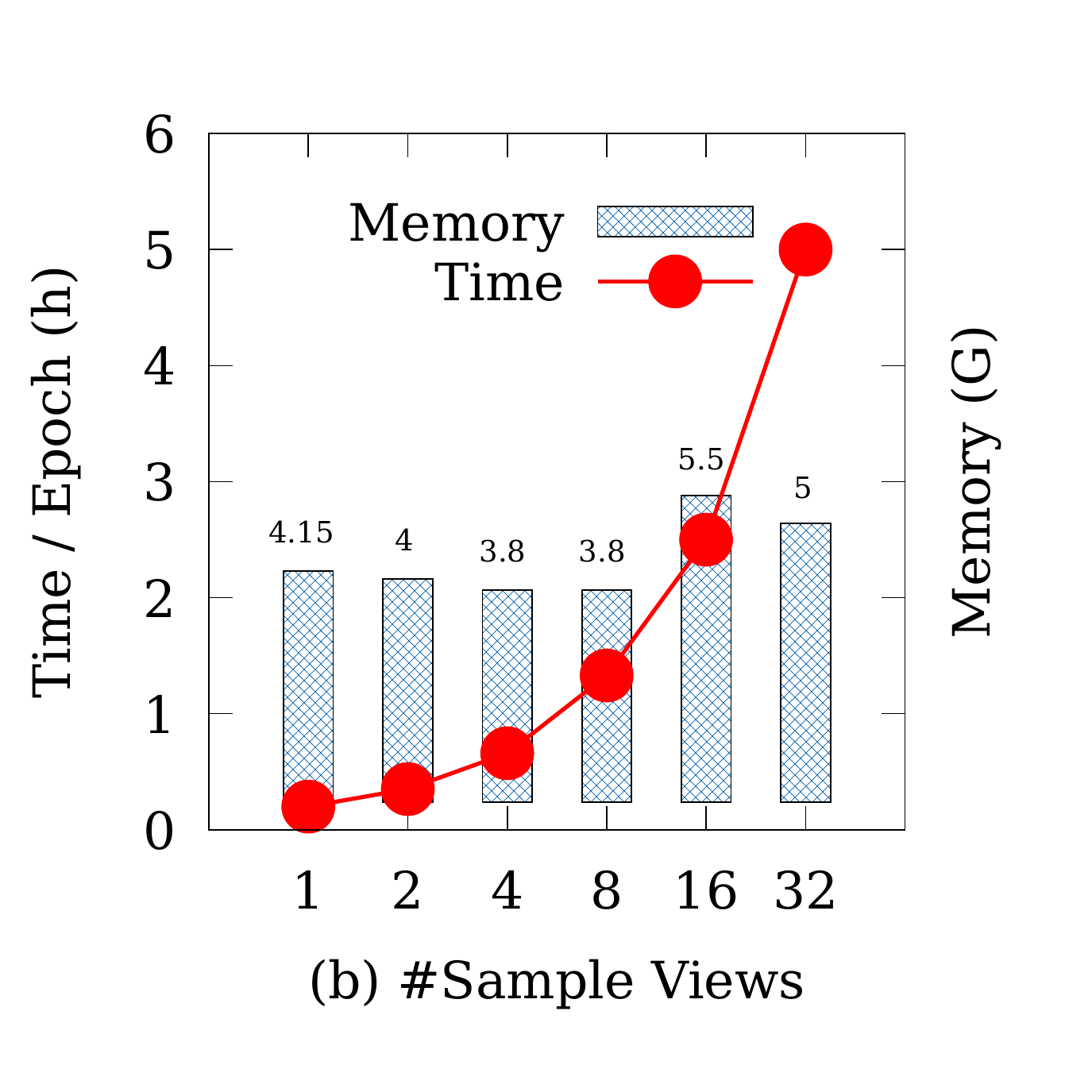}}
  \caption{Tuning of approximate contrastive training with different numbers of sampled views on Bio dataset. (a) Performance with different batch sizes. (b) Time/Memory Consumption with batch size $32$. Labels on the histogram indicate memory consumption.}
  \label{fig:sample}
\end{figure}

\subsection{Training Strategy Analysis}\label{sec:exp-training}\hfill\\
As mentioned above, we propose an approximate contrastive training strategy to reduce the time and space consumption of GMPT-CL and the corresponding theoretical analysis about time and memory complexity  can be found in In Section~\ref{sec:appro}.
Here we conduct detailed experiments to show how the fine-tuned GNNs' performance is affected by the batch size $n$ and number of sampled views $q$,
and what's the actual running time/memory when the proposed approximate training strategy is applied.

\subsubsection{Performance Comparison \emph{w.r.t.} Batch Size and the Number of Sampled Views}
We pre-train GIN models in different batch sizes $n\in\{4,8,16\}$ and numbers of sampled views $q\in\{1,2,4,8,16,32\}$,
and compare the fine-tuned model's performance on downstream dataset of Bio.
Note that for a certain batch size $n$, the maximum number of sampled views is $q=2n$ (totally $2n$ views).

As Figure~\ref{fig:sample:auc} shows, with different batch sizes and number of sampled views,
our method can reach no less than $69\%$ ROCAUC performance.
We notice that applying approximate contrastive training usually improves the performance of GMPT-CL.
Generally, GMPT-CL reaches the best performance when the number of sampled views $q=4$.
We speculate that the performance gain comes from the randomness introduced by contrastive approximate training.
Besides, we find that even with a small sampling number $q=1$, our method can still have competitive performances.

\subsubsection{Time and Memory Consumption \emph{w.r.t.} the Number of Sampled Views}
Here, we take an intuitive look at the actual time and space consumption of the proposed approximate contrastive training strategy.
As Figure~\ref{fig:sample:timespace} shows, we find that as $q$ decreases, the training time of GMPT-CL per epoch also decreases,
which verifies that choosing a relatively small $q$ can dramatically reduce the training time of GMPT-CL.
Besides, we find that the memory consumption doesn't change a lot with different choices of $q$, verifying that the space complexity doesn't affect much by $q$.

In summary, we suggest adopting the proposed approximate contrastive training strategy when pre-training GNNs with GMPT-CL,
which has been shown to reduce the time/space consumption and gain a slight performance increasing.
For the choice of the number of sampled views $q$ with the batch size $n$,
we suggest to select a relative small $q$ (i.e., $4 \leq q \leq n$).

\begin{figure}
  \setlength{\belowcaptionskip}{-0.5cm}
  \centering
  \subfigure{\label{fig:transfer:l2p}\includegraphics[width=0.24\textwidth]{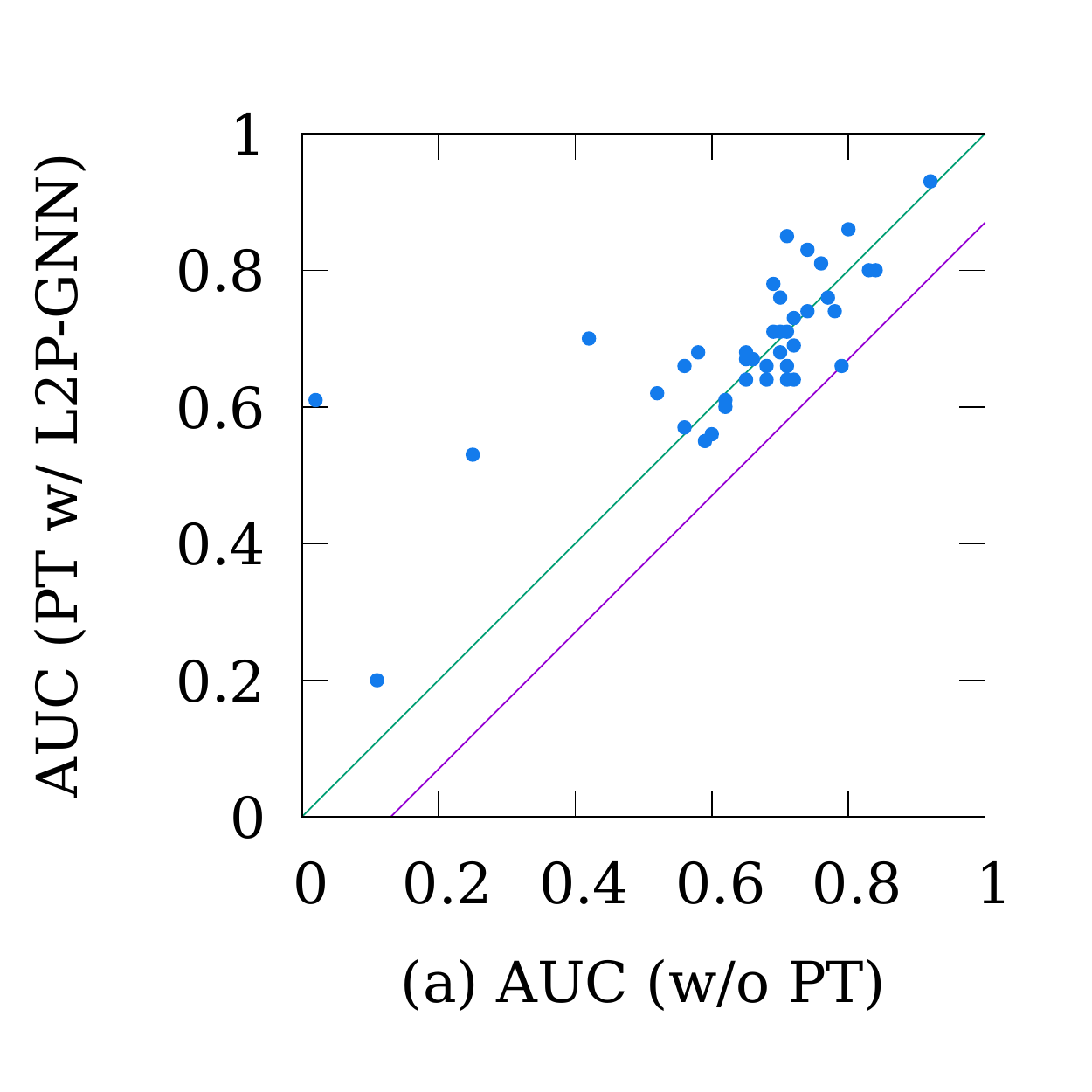}}
  \subfigure{\label{fig:transfer:gmcl}\includegraphics[width=0.24\textwidth]{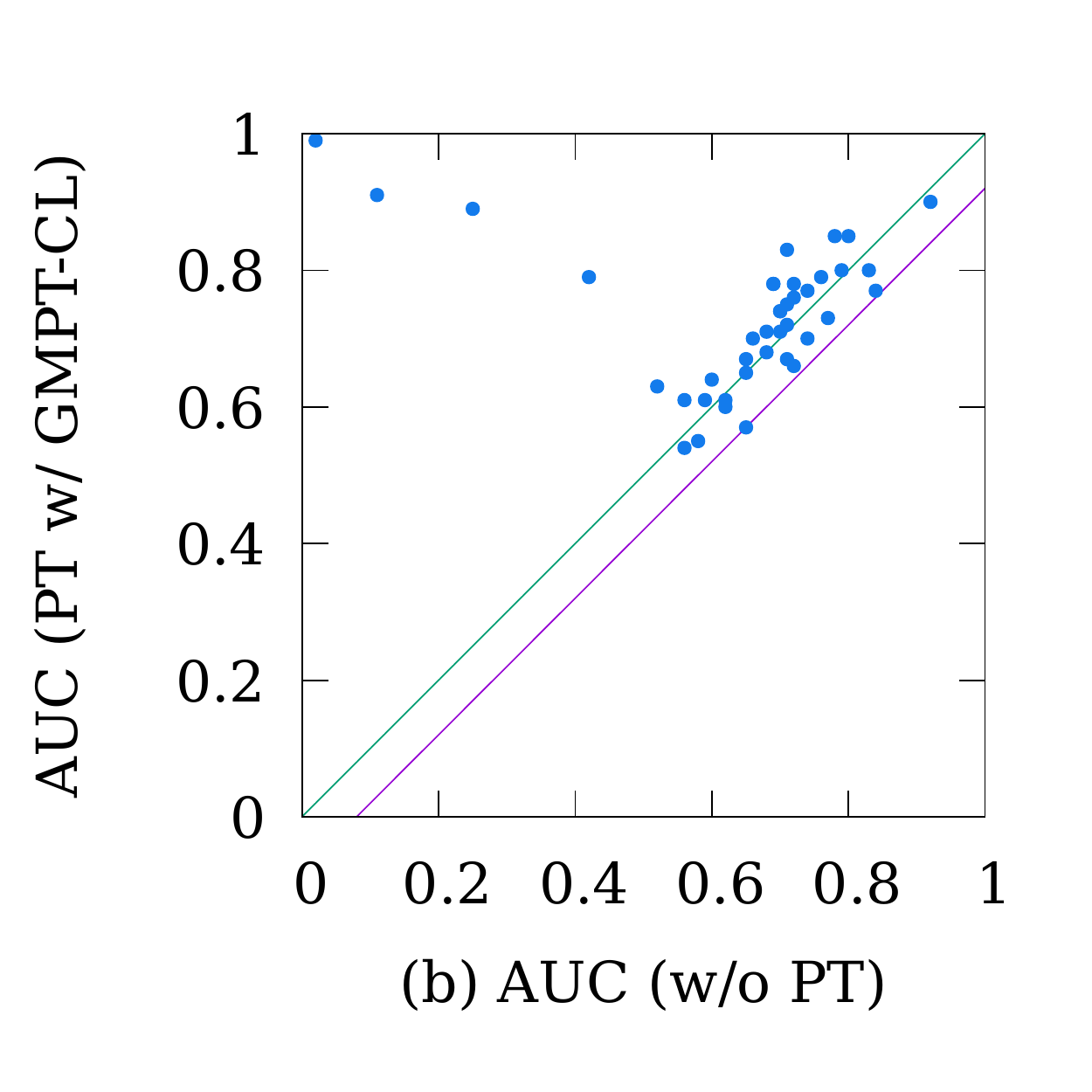}}
  \caption{Analysis of transferability for (a) L2P-GNN and (b) GMPT-CL over $40$ out-of-distribution subtasks of Bio. ``PT'' denotes as ``pre-training''. The green line indicates the borderline. The purple line indicates the worst negative transfer across the 40 subtasks.}
  \label{fig:transfer}
\end{figure}

\subsection{Transferability Analysis}\hfill\\
Out-of-distribution (OOD) problem widely exists in real-world applications,
meaning that graphs in the training set are structurally very different from graphs in the test set~\cite{hu2020strategies}.
Existing study shows that improperly designed GNN pre-training tasks may cause serious negative transfer.
Thus, we analyze the proposed GMPT-CL and the best baseline L2P-GNN to analyze the transfer status over the individual subtasks of out-of-distribution datasets.
The left and up area indicates positive transfer, and the right and bottom area indicates negative transfer.
As shown in Figure~\ref{fig:transfer}, we can see that compared to the best baseline L2P-GNN,
the proposed GMPT-CL has less negative transfer cases ($12$ \emph{v.s.} $17$),
as well as a slighter negative transfer extent ($-0.07$ \emph{v.s.} $-0.13$).
Besides we can see that GNN pre-trained by GMPT-CL gets AUC result $>0.5$ on all downstream subtasks of Bio.
All the observations above show the good transferability of the proposed GMPT-CL method.

\section{Conclusion}
In this work, we propose GMPT, a general graph matching-based GNN pre-training framework for both self-supervised pre-training and coarse-grained supervised pre-training.
By structuralized neural graph matching module,
we generate adaptive representations for the matched graphs,
encouraging GNNs to learn both globalized and localized domain-specific semantics in a single pre-training task.
We also propose approximate contrastive training strategy,
which significantly reduces the time/memory consumption brought by the graph matching module.
Extensive experiments on multi-domain out-of-distribution benchmarks show the effectiveness and transferability of our method.

Besides GMPT,
more hybrid-level GNN pre-training tasks can be explored in the future.
In addition, we will also consider generalizing our framework to more complicated graph structures (e.g. dynamic graphs, knowledge graphs, and heterogeneous graphs).

\section*{Acknowledgement}
This work was partially supported by the National Natural Science Foundation of China under Grant No. 61872369 and 61832017,
Beijing Outstanding Young Scientist Program under Grant No. BJJWZYJH012019100020098,
and CCF-Ant Group Research Fund.
Xin Zhao is the corresponding author.

\bibliographystyle{siam}
\bibliography{main}

\end{document}